# Decision-Theoretic Troubleshooting: A Framework for Repair and Experiment

**John S. Breese    David Heckerman**

Microsoft Research
One Microsoft Way
Redmond, WA 98052-6399
{breese,heckerma}@microsoft.com

## Abstract

We develop and extend existing decision-theoretic methods for troubleshooting a nonfunctioning device. Traditionally, diagnosis with Bayesian networks has focused on belief updating—determining the probabilities of various faults given current observations. In this paper, we extend this paradigm to include taking actions. In particular, we consider three classes of actions: (1) we can make observations regarding the behavior of a device and infer likely faults as in traditional diagnosis, (2) we can repair a component and then observe the behavior of the device to infer likely faults, and (3) we can change the configuration of the device, observe its new behavior, and infer the likelihood of faults. Analysis of latter two classes of troubleshooting actions requires incorporating notions of persistence into the belief-network formalism used for probabilistic inference.

## 1   Introduction

Given that a device is not working properly or a patient has some complaint, automated-diagnostic systems have traditionally been been designed to determine the set of faults or diseases that explain the symptoms [de Kleer and Williams, 1987, Genesereth, 1984, Heckerman et al., 1992, Breese et al., 1992]. The diagnostician is able to ask questions about the behavior of the device or test individual components in order to determine if they are working properly. As new information is gained, the procedure updates its current view of the world. Inference focuses on identifying the set of faults consistent with the observations and, in the probabilistic case, assigning probabilities to the feasible diagnoses. Information gathering proceeds until a single cause has been identified or the current diagnosis is sufficiently restricted to support action.

From a decision making perspective, however, our primary objective is to repair the device or cure the patient, not just determine what is wrong. At any stage of the process, there are many possible observations, tests, or repairs that can be applied. Because these operations are expensive in terms of time and/or money, we wish to generate a sequence of actions that minimizes costs and results in a functioning device (or healthy patient). In this paper, we develop a diagnostic procedure (i.e., planner) that selects the next best action by estimating the expected cost of repair for various plans. Because computation of the optimal plan is intractable, we develop procedures for estimating the expected cost of repair for plans that

1. Repair system components in sequence (with no observations) until the device is repaired

2. Gather evidence about the state of the system and then repair components

3. Change the configuration of the system, observe behavior, and then repair components

We use these cost estimates myopically to determine the next troubleshooting step. After each action, probabilities are updated and a new set of potential plans are generated. This cycle continues until the device is working properly.

From the perspective of belief-network diagnosis, the second two classes of troubleshooting action are problematic. When we actively set (as opposed to passively observe) the value of a variable in a Bayesian network, previous observations regarding device behavior may change. For example, in Figure 1 depicting a simplified model of printing for personal computers, we may initially be unable to print over the network, but we are able to print after we set the print logic to print locally.

There is no mechanism in a single static Bayesian network to combine the observations before and after this action. In this paper, we describe a notion called *persistence*, and show how it can be used as a solution to this problem in the context of troubleshooting.

## 2 Basic Troubleshooting

In this section, we describe a set of assumptions under which it is possible to identify an optimal sequence of observations and repair actions in time proportional to the number of components in the device, without explicitly constructing and rolling back a decision tree. The approach is described in Heckerman et al. (1995). Let us suppose that the device has $n$ components $c_1, \ldots, c_n$ and each component is in exactly one of a finite set of states. We assume[1]

1. There is only one problem-defining variable in the Bayesian network for the device. This variable represents the functional status of the device. One of the states of this variable must correspond to normal operation. In Figure 1, the node labeled "Printer Output" is the problem-defining node. We will write $e = normal$ to denote the event that the problem defining node is normal.

2. At the onset of troubleshooting, the device is faulty—that is the problem defining variable is observed to be faulty.

3. Single fault: Exactly one component is abnormal and is responsible for the failure of the device. We use $p_i$ to denote the probability that repairing component $c_i$ will repair the device under the current state of information I, that is $p_i \equiv \Pr(e = normal|repair(c_I), \text{I})$. Under the single-fault assumption, $\sum_{i=1}^{n} p_i = 1$.

4. Each component is observable or unobservable. An observable component can be unambiguously tested or inspected to determine correct operation. In this formulation, an observable component that is observed to be abnormal is immediately repaired. An unobservable component can never be directly observed, but can be repaired or replaced. In Figure 1, the components "Cable Port Hardware", "Network Connection", and "Driver File Status" are unobservable—one can reinstall the driver file from disk, but it is extremely difficult to determine if the file is corrupt directly. For convenience, we use *observation–repair action* to refer both to the observation and possible repair of an observable component and to the repair of an unobservable component.

6. The costs of observation and repair of any component do not depend on previous repair or observation actions.

7. Limited observations: No observations, other than the observation-repair of a component, are undertaken during the course of troubleshooting.

For the moment, let us consider only observable components. Let $C_i^o$ and $C_i^r$ denote the cost of observation and repair of component $c_i$. If we observe and possibly repair components in the order $c_1, \ldots, c_n$, then for the expected cost of repair under state of information I, denoted ECR(I), we have

$$\begin{aligned}
\text{ECR(I)} &= (C_1^o + p_1 C_1^r) + \\
&\quad (1 - p_1)(C_2^o + \frac{p_2}{1 - p_1} C_2^r) + \\
&\quad (1 - p_1 - p_2)(C_3^o + \frac{p_3}{1 - p_1 - p_2} C_3^r) + \cdots \\
&= \sum_{i=1}^{n} \left[ \left(1 - \sum_{j=1}^{i-1} p_j\right) C_i^o + p_i C_i^r \right]
\end{aligned}$$

That is, we first observe component $c_1$ incurring cost $C_1^o$. With probability $p_1$, we find that the component is faulty and repair it (and the device) incurring cost $C_1^r$. With probability $1 - p_1$, we find that the component is functioning properly, and observe component $c_2$. With probability $p_2/(1 - p_1)$, we find that $c_2$ is faulty and repair it; and so on.

By reversing the order of the any two steps in the repair sequence, one can show that the minimum expected cost sequence is obtained by ordering the components according to a decreasing probability to cost ratio. The expected cost of repair $ECR(\text{I})$ is understood to be that under the component ordering where

$$\frac{p_1}{C_1^o} \geq \frac{p_2}{C_2^o} \geq \cdots \geq \frac{p_n}{C_n^o}.$$

We sequentially repair or replace the components according to the ordering of their $p_i/C_i^o$ ratios. After each repair, we observe whether the device is working properly, and if so, terminate.

Including unobservable components in this approach is straightforward. Recall that an unobservable component $c_i$ is simply repaired with cost $C_i^r$. Therefore, an unobservable component acts just like an observable component that is observed with cost $C_i^r$ and always found to be faulty and repaired with cost zero. Consequently, we can include unobservable components in our procedure, provided we set $C_i^o$ to the cost of repair of unobservable component $c_i$, and set $C_i^r$ to zero.[2]

---

[1] The appropriateness of these assumptions is discussed in Heckerman et al. (1995).

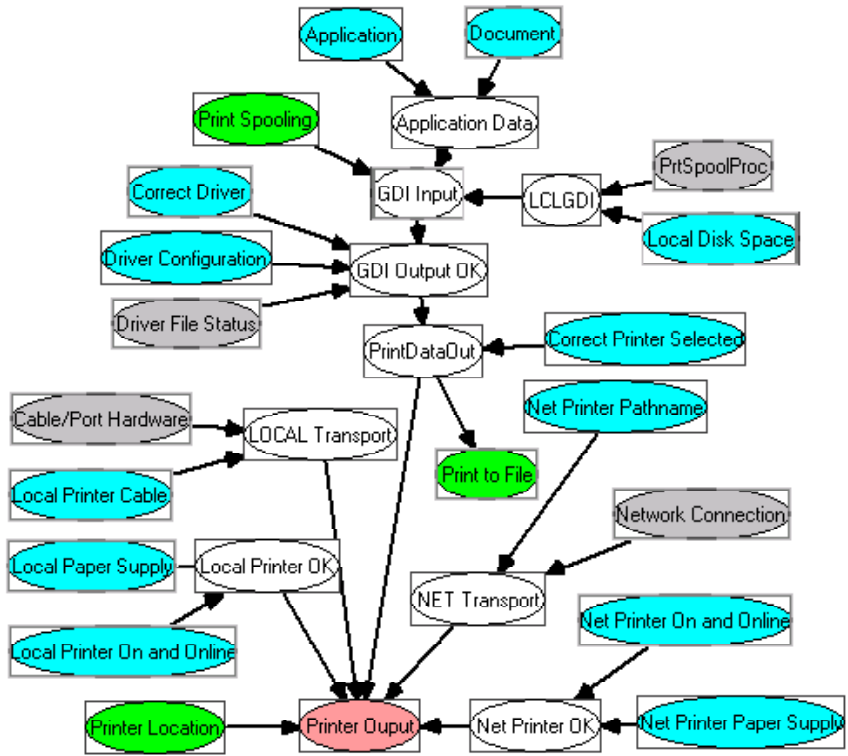

Figure 1: A simplified Bayesian network for printing problems on personal computers.

## 2.1 Computing Probabilities of Faulty Components using Persistence

When troubleshooting under uncertainty, we need to compute the probabilities that components have failed. In our approach, we compute these probabilities using a *Bayesian network*. Given the observation that "Print Output" is false, we can use a Bayesian-network inference algorithm to compute the probability that any or all of the system components are faulty. When we wish to recalculate probabilities given a component of the device has been repaired, updating procedures must account for the change in underlying state of the device and the fact that previous observations may have been invalidated.

Consider the simple causal relationship between the status of a computer's connection to the network ("Net"), which we model as having states `normal` or `abnormal`, and the appearance of the printer icon in the print manager ("Icon"), which we model as having states `normal` and `grey`. The two nodes are dependent as depicted in the Bayesian network in Figure 2a.

---
[2] It is also straightforward to add the notion of a service call, a fixed charge that will repair any fault with the device. See Heckerman et al. (1995) for details.

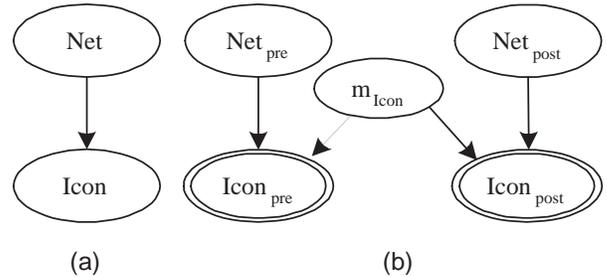

Figure 2: (a) A Bayesian network for the interaction between net connections and print icon appearance. (b) A persistence network for determining the probability that the print icon will appear normal after we verify net connectivity, given that the icon is currently greyed out.

Now suppose we observe the icon to be grey, and we want to determine the probability that the icon will be normal, after we make sure the network is connected. We can do so, using the Bayesian network shown in Figure 2b. In this network, "Net$_{pre}$" and "Net$_{post}$" represent whether or not the network is connected, re-spectively before and after we establish connectivity. Similarly, "Icon$_{pre}$" and "Icon$_{post}$" represent whether or not the icon is normal before and after we connect the network, respectively. The node $m_{Icon}$ represents all of the possible mappings between "Net" and "Icon". Following Heckerman and Shachter (1995), we call $m_{Icon}$ and its corresponding variable a *mapping node* and *mapping variable*, respectively. We call the remaining nodes and their corresponding variables *domain nodes* and *domain variables*, respectively.

As shown in Table 1, the mapping node $m_{Icon}$ has four possible states: (1) `ok`, where the icon is grey if and only if there is no network, (2) `stuck on grey`, where the icon is grey regardless of state of the network, (3) `stuck on normal`, where the icon is always normal, and (4) `backwards`, where the icon reads grey if and only if the network is connected. The node "Icon$_{pre}$" is a deterministic function of its cause "Net$_{pre}$" and the node $m_{Icon}$, as indicated by the double ovals around the node "Icon$_{pre}$". For example, if "Net$_{pre}$" is abnormal and $m_{Icon}$ is `backwards`, then "Icon$_{pre}$" will be `normal`. The node "Icon$_{post}$" is the same deterministic function of cause "Net$_{post}$" and the node $m_{Icon}$. The uncertainty in the relationship between "Net" and "Icon" is encoded in the probabilities for the node $m_{Icon}$. These probabilities are constrained by, but not necessarily determined by, the probabilities in the Bayesian network.

By using a single node $m_{Icon}$ to represent the mappings between "Network" and "Icon" both before and after the action is taken, we encode the assertion that the mapping (or mechanism) between cause and effect is not affected by actions that may change the cause. Although we are uncertain about which mapping holds, representing this uncertainty in a single node enforces the assertion that the mapping persists across the action. We could equivalently have two mapping nodes, one for before and one for after the action, with an arc between the two. The state transition distribution for the mapping variables in this case would encode the restriction that the state of the mapping variable after the action is the same as the state before the action.

In order to calculate the probability that the icon will be normal after establishing network connectivity, we set "Net$_{post}$" to state `normal` and set "Icon$_{pre}$" to state `grey` in the network in Figure 2b. We then apply a standard belief network inference algorithm to to the network to calculate the probability that "Icon$_{post}$" = `normal`.

We refer to this notion as *causal persistence*. Causal persistence is closely related to Heckerman and Shachter's (1995) concept of *unresponsiveness*. The latter notion, however, pertains to transitions from some state of the world to a counterfactual state of the world, whereas causal persistence pertains to transitions between real states of the world at different times. The notion of causal persistence (although not in this terminology) is discussed in Heckerman et al. (1994). Related notions are discussed in Pearl (1993), Goldszmidt and Darwiche (1994), and Balke and Pearl (1994).

In general, suppose we have a Bayesian network for a set of domain variables $\mathcal{U} = \{x_1, \ldots, x_n\}$. Further, suppose that we want to answer questions of the form: "What would be the probability of $\mathcal{X} \subseteq \mathcal{U}$ if we were to take some action, given that we now observe $\mathcal{Y} \subseteq \mathcal{U}$." To answer such questions, we construct a new Bayesian network as we did in our example. First, we copy the network, using the first and second instances of the network to represent the domain variables before and after we take the action. Second, we introduce mapping nodes $m_i$ for each domain node $x_i$, and assess the prior probabilities of these nodes. Typically, these mapping nodes will be mutually independent, but they need not be [Heckerman and Shachter, 1995]. Also, note that if $x_i$ is a root node, then $m_i = x_i$ and we do not need to explicitly create the mapping node. Third, we make both versions of node $x_i$ the same deterministic function of its parents and $m_i$. Finally, we identify those domain nodes in the post-action network that are affected directly by our action, break the arcs from their parents, and set the states of these nodes to their values as determined by our action.[3] Under this construction, as in our example, the shared mapping nodes encode causal persistence. We refer to Bayesian networks constructed in this manner as *persistence networks*.

## 2.2 Single-Copy Approximation: Repair

Although a persistence network is a correct representation of the effects of actions, we are still left with the problem that a persistence network often contains many undirected cycles, making inference computationally expensive. In this section, we describe an approximation wherein the repair probabilities can be computed without copying the original Bayesian network. This approximation relies on representing the behavior of the device using a form of causal independence (see, e.g., Srinivas [1993] and Heckerman and Breese [1996]).

The notion of causal independence is illustrated in Figure 3. The effect of set of causes $c_1, \ldots, c_n$ on an effect

---

[3]In doing so, we assume that the direct effects of actions are deterministic. The more general case can be handled with influence diagrams in canonical form [Heckerman and Shachter, 1995].

Table 1: The four possible mappings between "Net" and "Icon".

|      | ok       |        | stuck on grey |        | stuck on normal |        | backwards |        |
|------|----------|--------|---------------|--------|-----------------|--------|-----------|--------|
| Net  | abnormal | normal | abnormal      | normal | abnormal        | normal | abnormal  | normal |
| Icon | grey     | normal | grey          | grey   | normal          | normal | normal    | grey   |

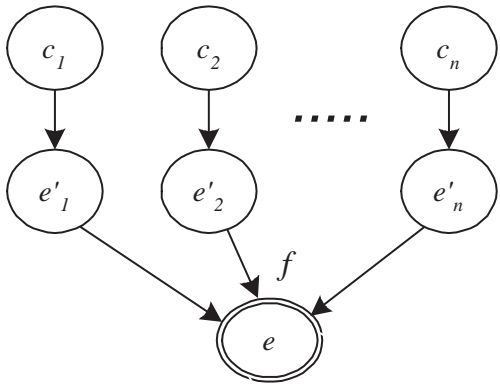

Figure 3: A Bayesian network for the interaction between a set of causes and an effect under causal independence.

$e$ are modeled in terms of a set of *mediator* variables $e'_1, \ldots, e'_n$ and a function $f(e'_1, \ldots, e'_n)$. The mediator variables are dependent on their associated causes $c_i$, and serve as inputs to the function $f$. The choice of $f$ is arbitrary, in general, though in many applications it is an *or* relationship, where the effect is *abnormal* if any of the causal inputs is *abnormal*. Typically, we also designate one of the causal inputs as a "leak" cause that is always set to its abnormal state. Note that, for purposes of troubleshooting as described in Section 2, the leak causes are summarized as a single "all other" fault that can be repaired at a fixed cost with a service call, as discussed in [Heckerman et al., 1995].

Suppose our device satisfies the following conditions: (1) the interaction between the $n$ components of the device $c_1, \ldots, c_n$ and the output of the device $e$ satisfy causal independence, and (2) the function $f$ has the property that the output is normal if all inputs are normal. For example, the noisy-or relationship and the noisy-adder relationship (with 0 corresponding to `Normal`) satisfy these conditions. If we impose these restrictions, and we assume there is a single component that is faulty, Heckerman et al. (1994) show that

$$\Pr(e_{post} = \texttt{Normal}|Repair(c_i), e_{pre} = Abnormal, \texttt{I})$$
$$= \Pr(e'_{i,pre} = Abnormal|e_{pre} = Abnormal, \texttt{I})$$

That is, the probability that the effect is normal after the repair is equal to the probability that the associated mediator node is abnormal in the original network. For many troubleshooting domains, we have used the right-hand-side of Equation 1 in place of the left-hand-side as an approximation, because the right-hand-side may be computed without copying the Bayesian network for a device. Given the observation that $e$ is abnormal, the single-fault assumption is likely to be true, because it is unlikely that two components of a device will fail at the same time. We rescale the probabilities for the nodes $e'_i$, so that the sum over the probabilities of all non-normal states of all nodes $e'_i$ is equal to one.

### 2.3 Nonbase Observations

In Section 2, we considered two special classes of observations: (1) the observation of the problem-defining variable after a repair is made, and (2) the observation of a component before a repair is made (as part of an observation-repair action). We refer to these observations as *base observations*. In many situations, we want to be able to make more general observations. For example, when our car fails to start, we may want to check the radio or the headlights in order to check the status of the electrical system. In this section, we describe a method for making such general observations.

Let us suppose we have $m$ nonbase observations $o_1, o_2, \ldots, o_m$ available to us. We assume that observation $o_i$ can take on exactly one of $r_i$ possible states. We write $o_i = k$ to indicate that observation $o_i$ takes on state $k$. First, we use the procedures described in Sections 2 to generate a troubleshooting sequence consisting of only base observations and repairs. Second, we imagine that we make observation $o_i$ first, and then determine the sequence. The expected cost of observing $o_i$ with information $\texttt{I}$, denoted $\text{ECO}(\texttt{I}, o_i)$, consists of an observation cost plus the expected cost of a set of conditional plans as given by

$$\text{ECO}(\texttt{I}, o_i) = C_i^o + \sum_{k=1}^{r_i} \Pr(o_i = k|\texttt{I})\ ECR(\texttt{I} \cup \{o_i = k\}) \quad (1)$$

Note that the troubleshooting sequence following the observation may be different for every possible out-come of the observation. Finally, we repeat the computation of ECO for every possible nonbase observation.

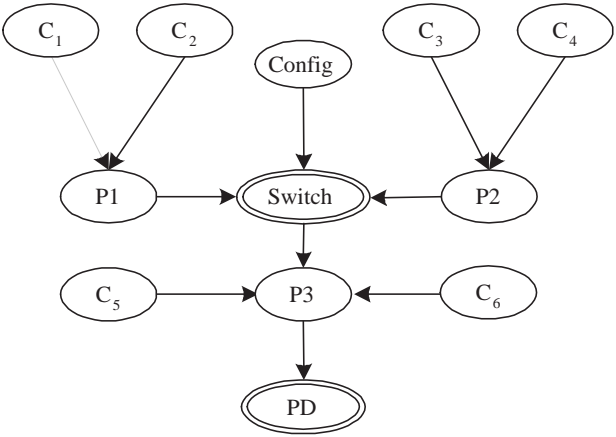

Figure 4: A Bayesian network for a simple configuration scenario.

## 3 Changing the Configuration

An expert troubleshooter will often change the setup or configuration of a device and observe the new behavior of the device to infer the underlying problem(s) with the device. In this section we provide a decision theoretic account of this type of troubleshooting, and show how it can be implemented using Bayesian networks and the troubleshooting framework we have described above.

A Bayesian network for a canonical configurable device is shown in Figure 4. The device works with either the $P1$ process or the $P2$ process. The node *Config* corresponds to an action that determines the configuration of the device. The node *Switch* takes on the value of $P1$ if *Config* is set to 1 and the value of $P2$ if *Config* is set to 2. The overall behavior of the device, captured in the problem defining node ($PD$), depends on *Switch* and two additional subcomponents, $C_5$ and $C_6$. By setting the configuration node to one value or the other, we isolate part of the system for inspection.

To further motivate the scenario, consider a failure in printing from your PC to a network printer. In general, the problem could be local to your PC or somewhere in the network. We can try to print locally. If this action succeeds, then it rules out a host of problems associated with the local configuration. If this action fails, it will increase the probability of a local malfunction. Though changing configuration appears to be most compelling in troubleshooting hardware or other man-made devices, the notion is also relevant for medical diagnosis. For example, a physician diagnosing headaches may ask a patient to change diet or work habits to rule out various allergy or stress related disorders.

To incorporate this type of reasoning into our framework, we need two elements. First, we need to calculate the updated fault probabilities, given we set a configuration parameter and then make an observation. Because we are combining information from two situations—the first with the configuration parameter set in its original position and the second after the change—we will use a persistence network. Second, we need to use these updated fault probabilities to estimate the expected cost of a plan that starts with a configuration change. We will describe the cost estimation framework first, and then discuss calculation of the necessary probabilities.

Our framework is as follows. In most cases, a configuration node can be treated like any other non-base observation (see Section 2.3) in that we may recommend that the current state be observed without necessarily setting it to another state. In addition, we allow the following new alternative: (1) set configuration node $c$ to some value $m$, (2) make a non-base observation $o_i$, (3) change the configuration back to its original state, and (4) undertake a sequence of repairs. The analysis of this new alternative is similar to that described in Section 2.3. The expected cost of repair for setting configuration node $c$ to state $m$ followed by observing node $o_i$ is given by

$$\text{ECCO}(\texttt{I}, o_i, c = m) = C^s + C^o_i + \sum_{k=1}^{r_i} \Pr(o_{i,post} = k | c_{post} = m, \texttt{I}_{pre}) \\ ECR(\texttt{I}_{pre} \cup \{o_{i,post} = k\} \cup \{c_{post} = m\}) \quad (2)$$

where $C^s$ is the cost of changing the configuration from its current state to a new state and back. In the course of troubleshooting, we evaluate this expression for each possible configuration–observation pair,[4] and recommend that configuration–observation pair that has the lowest expected cost.

### 3.1 Calculating Probabilities after a Configuration Change

As mentioned, we need to calculate the probability of the faults given we have set a configuration variable to some value and then made some observation. We use a persistence network to perform this computation. As discussed in Section 2.2, we will again rely on causal independence as a representation of the device's operation. Here the motivation is not to allow a single-copy approximation, but to reduce the state space of the mapping nodes needed in the persistence network. Causal independence imposes a special structure on the mapping nodes associated with an effect node. Specifically, there is a single mapping node associated with each cause to mediator link and these mapping nodes are mutually independent.

---
[4]The observation can be restricted to the problem defining node, if desired.

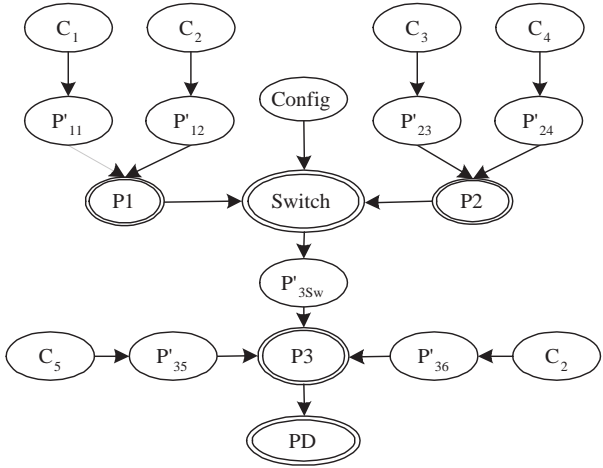

Figure 5: A Bayesian network for a simple configuration scenario, expanded to expose its causal independence representation.

The first step is to convert the original Bayesian network into a causal-independence representation with mediator nodes. This conversion was applied to the network of Figure 4 and is shown in Figure 5. We then create mapping nodes for each non-root non-deterministic domain variable as shown in the upper left-hand-corner of Figure 6. We then copy all nodes in the original network—except the component nodes—and include a new node that represents the configuration after the change. We do not copy the component nodes, because we assume their states persist after the configuration change. The result is shown in Figure 6. Finally, we copy the parent-structure and probability/function tables of the copied nodes to their corresponding copies.

The network in Figure 6 is used to evaluate the probabilities used in Equation 2. We retain all current observations in the original (upper left) portion of the network, presumably including the fact that the problem defining node is abnormal. We set the configuration node in the "post" network to a possible value, and perform probabilistic inference using the network to obtain $\Pr(o_{i,post} = k | c_{j,post} = m, \mathtt{I}_{pre})$ for each possible observation $o_{i,post} = k$. We then compute the fault probabilities of the components for every possible non-base observation. We use these probabilities to calculate the various repair sequences and associated expected costs of repair as described previously. Finally, we repeat this entire set of computations for every possible state of the configuration node other than its current state. Note that if there are multiple configuration nodes and we hypothetically set the value of one configuration node, the value of the other configuration nodes must remain the same from the "pre" to "post" network, even if we do not know their values.

### 3.2 Single-Copy Approximation: Configuration

An approximation can be undertaken to avoid the use of the persistence network for the computation of the fault probabilities. First, we set the configuration node to some value and remove any down stream evidence in the original network. Next, for each possible non-base observation, we note which fault probabilities go to zero or one. Finally, we generate repair plans, clamping those logically determined component nodes to their respective values and retaining the original fault probabilities for the remaining component nodes.

## 4 Summary of the Troubleshooting Decision Making

In the course of executing this plan, we generate many complete repair plans, but at each stage we only execute the first step of one of the plans. Thus, our approximate decision-theoretic method for generating a troubleshooting sequence can be summarized as follows. First, we evaluate ECR(I)—the expected cost of repair under our current state of information I. Also, for every observation $o_i$, the expected cost of its observation ECO(I, $o_i$) is calculated. Also, for every combination of configuration setting and observation, we calculate ECCO(I, $o_i, c = m$). Next, we recommend that repair, observation, or configuration–observation that has the lowest expected cost. If ECR(I) has the lowest value, then we repair $c_1$, check if the device is functioning properly, and quit if so. If an observation or configuration–observation has the lowest expected value, we make that observation. Finally, we update probabilities in the network, based on the new information from the observation or the repair, and iterate this procedure.

Although we evaluate the expected cost of the next step by assuming that a non-base observation or configuration–observation will be followed by a pure repair sequence, in the actual running of the algorithm, observation and configuration–observations can be in-terleaved with the repair sequence. Similarly, although the cost-estimation step assumes a single fault, the algorithm will run (albeit approximately) for a device with multiple faults.

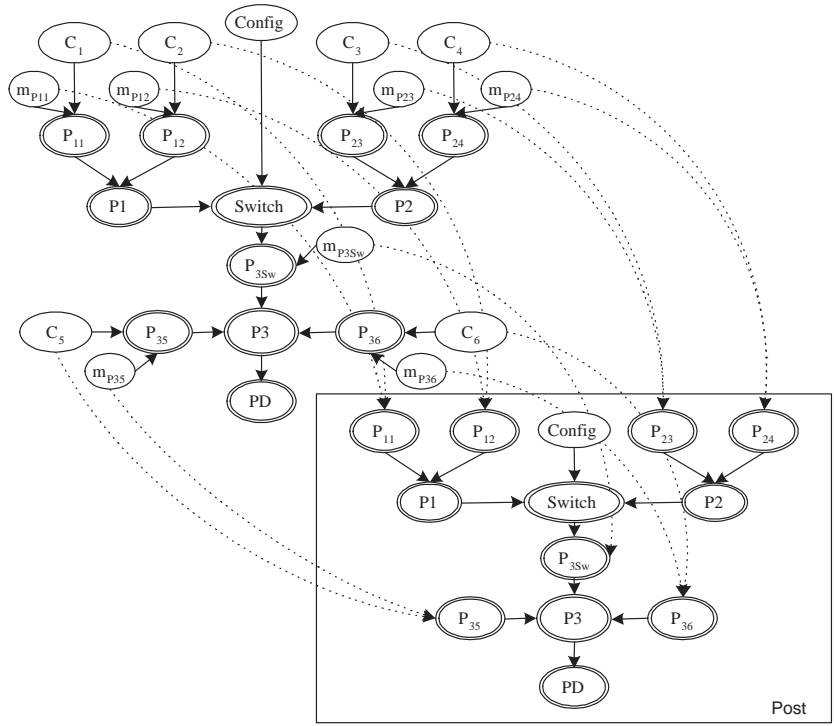

Figure 6: A persistence network for a simple configuration scenario.

## 5 Example

We have applied the general approach to troubleshooting printing problems, automobile startup problems, copier feeder systems, and gas turbines. A model addressing Microsoft Windows 95 printing problems, without the procedure for suggesting configuration changes described in Section 3, was implemented and delivered as part of the Microsoft Windows 95 Product Support Services Resource Kit. This system has been well received and additional troubleshooters based on this approach are in development.

An example of the type of steps generated by the troubleshooter follows. For the printer network, troubleshooting begins by observing the Print Output node in its abnormal state. A pure repair sequence—that is with no nonbase observations—starts with verifying that the network printer is on and online and has an expected cost of 13.62 minutes. The best nonbase observation is determination of the printer location, followed by a repair sequence, with a total expected cost of 13.71 minutes. The best configuration-observation option is to print locally and see if you get output. With probability 0.42 there is output and the expected cost of repair at that point is 7.07 minutes, with a sequence starting with verifying the net printer is online. Under this scenario, we have generally ruled out any problems with the computer itself. If local printing fails, (probability is 0.58), then a print driver problem (or some other local problem) is indicated and the expected cost of repair is 13.83 minutes. The expected cost of the configuration-observation plan is 10.98 minutes, plus 2 minutes for setting (and resetting) the configuration for a total of 12.98 minutes, the best option at this stage.

In terms of computational overhead, the primary factors are the increased size of the persistence network relative to the original network and additional inference steps needed for configuration recommendations. For the printing problems network (Figure 1) and the configuration network (Figure 4), the size of the join tree inference representation and propagation times for the persistence networks were roughly twice that for the standard single copy versions. We need a number of additional inference cycles equal to the product of the number of individual configuration settings and the number of individual observation values. The factors have not been significant in the examples we have run to date.

# 6  Empirical Results

In previous work [Heckerman et al., 1995], we developed a Monte-Carlo technique for estimating troubleshooting costs for a given planner and domain. We used a Bayesian network for a given device to generate a relatively large set of problem instances where one or more faults are known to have occurred. This study showed that the decision-theoretic troubleshooter had lower expected costs of repair than a static cost–based procedure, and also performed well in situations where there were multiple faults.

We performed a similar simulation study to verify the performance of the configuration planner versus a planner that did not suggest configuration changes. We applied each algorithm to the printing network shown in Figure 1. The planner that suggested configuration changes had an average time to resolution of 13.4 minutes, while the average time to repair for the planner that did not suggest configuration changes was 14.4 minutes. Of course, the magnitude of savings in a particular application depend on the cost and probabilities in that domain. For a domain such as printing problems where are there are millions of incidents per year, even the modest savings estimated here can result in a substantial total time or dollar savings.

# 7  Summary

We have developed a system for decision-theoretic troubleshooting where in a system can choose among several classes of possible actions—repairing a component, making a passive observation, or changing the configuration of the device and making an observation. We have developed approximations in the context of a myopic cycle for determining the best course of action. An important component of this work has been to show how to use the concept of persistence to compute the probabilities of events after repairs and configuration changes have been made.

## Acknowledgments

The authors thank Koos Rommelse and David Hovel for discussions and implementation related to this work.